\documentclass[letterpaper]{article} 
\usepackage[]{aaai2026}  
\usepackage{times}  
\usepackage{helvet}  
\usepackage{courier}  
\usepackage[hyphens]{url}  
\usepackage{graphicx} 
\usepackage{natbib}  
\usepackage{caption} 

\usepackage{url}            
\usepackage{booktabs}       
\usepackage{amsfonts}       
\usepackage{nicefrac}       
\usepackage{microtype}      
\usepackage{xcolor}         
\usepackage{enumitem}
\usepackage{amsmath,amssymb}
\usepackage{graphicx}
\usepackage{textcomp}
\usepackage{xcolor}
\usepackage{subcaption}
\usepackage{multirow}
\usepackage{tabularx} 
\usepackage{pifont}
\usepackage{colortbl}
\usepackage{pgfplots}
\usetikzlibrary{plotmarks}
\usepackage{tcolorbox}
\usepackage{soul}
\usepackage{fontawesome5}
\newcommand\scalemath[2]{\scalebox{#1}{\mbox{\ensuremath{\displaystyle #2}}}}
\usepackage{xspace}
\newcommand{\ie}{\textit{i.e.}, \xspace}      
\newcommand{\eg}{\textit{e.g.}, \xspace}      
\newcommand{\etc}{\textit{etc}. \xspace}      
\usepackage{algorithm}
\usepackage{algorithmic}
\usepackage[dvipsnames]{xcolor}           
\usetikzlibrary{plotmarks,shapes.geometric}
\usepackage{pgfplots}
\usepackage{pgfplotstable}
\usepackage{pgfkeys}
\usepgfplotslibrary{groupplots}
\usetikzlibrary{patterns, positioning}

\newcommand{\boldimgbox}[2]{%
  \begingroup
  \setlength{\fboxrule}{1.2pt}%
  \fcolorbox{#1}{white}{\includegraphics[height=1cm]{#2}}%
  \endgroup
}
\newcommand{\redimg}[1]{\boldimgbox{red}{#1}}
\newcommand{\greenimg}[1]{\boldimgbox{green}{#1}}

\usepgfplotslibrary{groupplots}

\newcommand{\modelname}{{DaSH}}

\newcommand{\dasflat}{DaS (flat)}

\definecolor{linkblue}{RGB}{0, 102, 204}

\usepackage{newfloat}
\usepackage{listings}
\DeclareCaptionStyle{ruled}{labelfont=normalfont,labelsep=colon,strut=off} 
\floatstyle{ruled}
\newfloat{listing}{tb}{lst}{}
\floatname{listing}{Listing}
\title{Hierarchical Dataset Selection for High-Quality Data Sharing}
\author {
    Xiaona Zhou\textsuperscript{\rm 1},
    Yingyan Zeng\textsuperscript{\rm 2},
    Ran Jin\textsuperscript{\rm 3},
    Ismini Lourentzou\textsuperscript{\rm 1},
}
\affiliations {
    \textsuperscript{\rm 1}University of Illinois Urbana-Champaign\\
    \textsuperscript{\rm 2}University of Cincinnati\\
    \textsuperscript{\rm 3}Virginia Polytechnic Institute and State University\\
    xiaonaz2@illinois.edu, zengyy@ucmail.uc.edu, jran5@vt.edu, lourent2@illinois.edu 
}

\begin{document}

\maketitle

\begin{abstract}
The success of modern machine learning hinges on access to high-quality training data. In many real-world scenarios, such as acquiring data from public repositories or sharing across institutions, data is naturally organized into discrete datasets that vary in relevance, quality, and utility. Selecting which repositories or institutions to search for useful datasets, and which datasets to incorporate into model training, are therefore critical decisions, yet most existing methods select individual samples and treat all data as equally relevant, ignoring differences between datasets and their sources.
In this work, we formalize the task of dataset selection: selecting entire datasets from a large, heterogeneous pool to improve downstream performance under resource constraints. We propose \ul{Da}taset \ul{S}election via \ul{H}ierarchies (\textbf{\modelname{}}), a dataset selection method that models utility at both dataset and group levels (\eg collections, institutions), enabling efficient generalization from limited observations.
Across two public benchmarks (\textsc{Digit-Five} and \textsc{DomainNet}), \modelname{} outperforms state-of-the-art data selection baselines by up to 26.2\% in accuracy, while requiring significantly fewer exploration steps. Ablations show \modelname{} is robust to low-resource settings and lack of relevant datasets, making it suitable for scalable and adaptive dataset selection in practical multi-source learning workflows. 

\end{abstract}

\begin{links}
\link{Project Page}{https://plan-lab.github.io/projects/dash}
\end{links}

\section{Introduction}\label{introduction}
Deep learning models have achieved impressive performance across a wide range of supervised learning tasks, largely due to their ability to leverage large, high-quality datasets~\cite{alzubaidi2023survey, sun2017revisiting, budach2022effects}. In many real-world scenarios, however, available data is distributed across multiple heterogeneous sources, such as publicly available dataset repositories or collaborating institutions, with varying degrees of relevance to a target task. A key challenge in such settings is determining which external datasets, if any, can meaningfully improve model performance~\cite{zhou2022domain, zhang2022survey}.

While practitioners often rely on intuition, domain expertise, or coarse metadata to guide dataset selection, there is little formal understanding of how to model such decisions algorithmically. 
Most existing approaches to data selection, \eg active learning~\cite{sener2018active, gal2017deep, christen2020informativeness, paul2017non, zeng2023ensemble}, data valuation~\cite{ghorbani2019data, pandl2021trustworthy, tang2021data, schoch2022cs, kwon2022beta}, \etc, operate at the instance level, selecting individual data samples and assuming that all datasets and data sources in the selection pool are uniformly relevant to the task. This assumption fails in multi-source settings, where data is naturally organized into datasets and repositories that vary in relevance, redundancy, and quality. In practice, datasets are typically acquired, licensed, or shared in discrete units, and often originate from common sources such as institutions, simulation pipelines, or web-scale repositories, which induce a hierarchical structure over the dataset pool. 

\begin{figure}[t!]
  \centering
  \includegraphics[width=0.91\columnwidth]{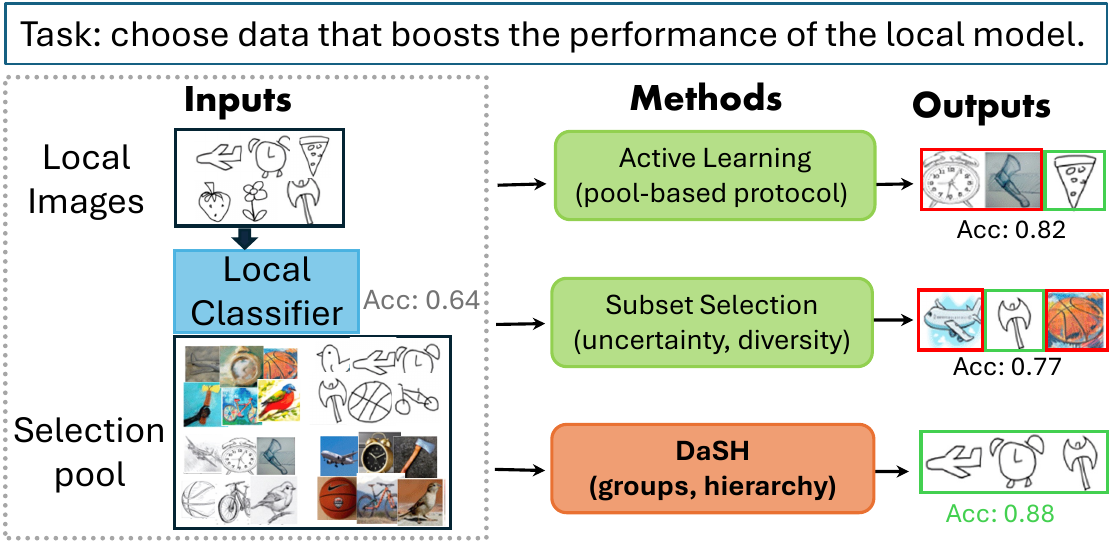} 
  \caption{Dataset selection aims to select entire datasets from external sources to improve local model performance. Instance-level methods, such as active learning and subset selection, ignore dataset structure and often select irrelevant or misleading samples. In contrast, \textbf{DaSH} leverages hierarchical grouping to efficiently identify relevant datasets, avoiding noisy sources and achieving higher downstream accuracy.}
  \label{fig:teaser}
\end{figure}
To address this gap, in this work, we formalize the task of \emph{dataset selection}: given a pool of datasets with unknown relevance to a target task, how can we efficiently identify a subset of datasets that will improve model performance, without having to exhaustively evaluate all candidates?
This setting, illustrated in Figure~\ref{fig:teaser}, reflects many real-world constraints, where data is acquired, licensed, or shared in dataset-level units and must be selected under resource, bandwidth, or labeling constraints from multiple sources such as web-scale repositories or partnering institutions.

To solve this new task, we propose \ul{Da}taset \ul{S}election via \ul{H}ierarchies (\textbf{\modelname{}}), a hierarchical Bayesian method that models dataset utility at both the group and dataset levels.
Given a large pool of candidate datasets, grouped based on dataset origin (\eg institution or collection), \modelname{} performs structured exploration to infer both group-level relevance and individual dataset utility via posterior inference over observed model performance. 
This hierarchical modeling allows \modelname{} to prioritize informative groups and avoid wasted evaluation on unrelated or harmful sources. Experiments on two benchmarks demonstrate \modelname{} significantly outperforms state-of-the-art baselines by up to 26.2\% in accuracy under low-resource settings.
The contributions of this work are:\looseness-1
\begin{itemize}[itemsep=0ex, parsep=0pt, topsep=-1.3pt, leftmargin=0.7cm]
    \item[\textbf{(1)}] We formalize the task of dataset selection from a heterogeneous pool of external datasets, a setting common in real-world workflows such as public data acquisition and cross-institutional collaboration, where data is organized into discrete, variably relevant sources.
    \item[\textbf{(2)}] We propose \modelname{}, the first dataset selection method that models dataset utility through hierarchical inference over groups and datasets, enabling efficient and robust selection under limited feedback.
    \item[\textbf{(3)}] We benchmark \modelname{} against four state-of-the-art data selection methods across two public datasets, demonstrating consistent performance gains, improving accuracy by up to 26.2\% \textsc{Digit-Five} and 10.8\% on \textsc{DomainNet}. Ablation studies show \modelname{} remains robust to grouping noise and scales effectively to large dataset pools, whereas existing methods frequently select irrelevant or low-utility data samples.
\end{itemize}

\section{Related Work} \label{sec:related_work}

\noindent \textbf{Data Selection.}
Improving model performance through strategic data selection has been extensively explored across various paradigms. In active learning, methods aim to minimize labeling costs by iteratively selecting the most informative unlabeled instances \cite{sener2018active, gal2017deep, christen2020informativeness, paul2017non, zeng2023ensemble, wang2023mutual, coleman2019selection}. Batch active learning extends this by selecting diverse subsets in each iteration to improve efficiency \cite{kirsch2019batchbald, kaushal2018learning}.
Beyond active learning, data valuation techniques assess the contribution of individual points to model performance. Approaches like Data Shapley \cite{ghorbani2019data} and its adaptations \cite{pandl2021trustworthy, tang2021data, schoch2022cs, kwon2022beta, liu20232d, courtnage2021shapley, wang2023data, just2023lava, yoon2020data, kwon2023data} quantify data utility, guiding the selection of valuable training instances. Additionally, subset selection methods \cite{killamsetty2021glister,coleman2019selection} focus on constructing representative subsets to expedite learning without compromising accuracy.

However, existing methods largely operate at the instance level and overlook the hierarchical structure often present in real-world settings, where datasets are naturally grouped into repositories, \eg by source or collection. In contrast, \modelname{} targets dataset selection, \ie identify groups of datasets that jointly maximize downstream performance. Empirical results demonstrate that incorporating hierarchical information improves selection efficiency and model robustness.

\noindent \textbf{Hierarchical Bandits.}
Hierarchical bandit algorithms address decision-making problems where actions are structured in a hierarchy, enabling efficient exploration and exploitation across multiple levels~\cite{hong2022hierarchical, munos2014bandits}. In recommendation systems, hierarchical bandits have been employed to model user preferences~\cite{yue2012hierarchical} and item categories~\cite{wang2018online, zuo2022hierarchical}, enabling personalized content delivery under resource constraints through adaptive frameworks~\cite{yang2020hierarchical, santana2020contextual}. Beyond recommendation, hierarchical bandits have been applied to intelligent tutoring, decentralized reinforcement learning, and multi-task off-policy learning \cite{castleman2024hierarchical, hong2023multi, kao2022decentralized}. These applications highlight the flexibility of hierarchical formulations in structuring complex decision processes across domains. Concurrently, theoretical advancements have focused on regret minimization and generalization across tasks using hierarchical Bayesian models \cite{kveton2021meta, hong2022hierarchical, guan2024improved}, offering principled frameworks for exploration under structured priors. 
Inspired by works in this space, our method tackles the unique setting of dataset selection by introducing a hierarchical Bayesian formulation that propagates dataset utility estimates across groups, enabling efficient amortization of training feedback via structured priors, and improving robustness to irrelevant or redundant datasets. 
To our knowledge, this is the first work to employ hierarchical bandits for dataset selection, with empirical evidence showing large gains in both accuracy and efficiency over non-hierarchical alternatives.

\section{Method}\label{sec:methods}

\noindent \textbf{Problem Definition.}
Consider $n$ data groups $\mathbf{g}\!=\!\{g_1, g_2, \dots, g_n\}\!=\!\{g_i\}_{i=1}^n$, where each group $g_i$ contains one or more datasets. Let the set of datasets in group $g_i$ be denoted $\mathbf{d}_i\!=\!\{d_{i,j}\}_{j=1}^{m_i}$, where $d_{i,j}$ is the $j$-th dataset in group $i$. Each dataset may contain an arbitrary number of data points. The full dataset pool is thus
\( 
\mathcal{D}\!=\!\bigcup_{i=1}^n \mathbf{d}_i\!=\!\{d_{i,j}\}.
\)
Given a local model $M_k$, the goal is to select a subset $\tilde{\mathcal{D}}_k \subseteq \mathcal{D}$ from external sources that maximizes the performance gain over training on the local data $d_k$ alone. Formally, we define:
\begin{equation}
\Delta \text{Acc}_k\!=\!\max_{\tilde{\mathcal{D}}_k \subseteq \mathcal{D}} \left( \text{Acc}(M_k, \tilde{\mathcal{D}}_k) - \text{Acc}(M_k, d_k) \right),
\label{eq:gain}
\end{equation}
where $Acc(M_k, d_k)$ is the performance of local model $M_k$ trained on local data $d_k$, $Acc(M_k, \tilde{\mathcal{D}}_k)$ is the performance of $M_k$ after training on selected datasets $\tilde{\mathcal{D}}_k$, and $\Delta Acc_k$ is the performance gain for model $\mathcal{M}_k$.  \looseness-1

\begin{figure*}[t!]
    \centering
    \includegraphics[width=\linewidth]{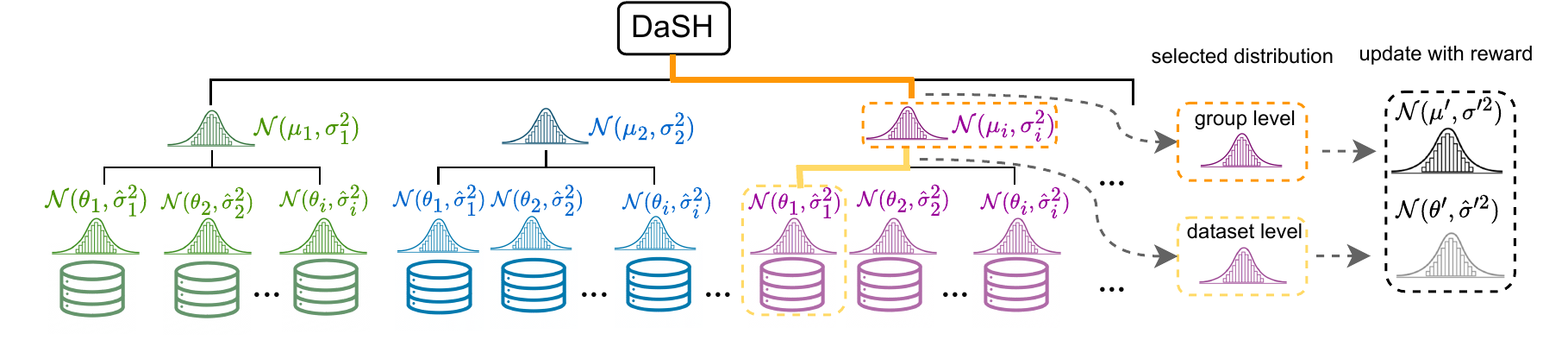}
    \caption{Overview of the \modelname{} dataset selection method. Each dataset and its corresponding group are modeled using Gaussian distributions $\mathcal{N}(\theta_{i}, \hat{\sigma}_i^2)$ and $\mathcal{N}(\mu_i, \sigma_i^2)$ for datasets and dataset groups, respectively. 
    The selection process involves choosing a dataset group, followed by a specific dataset within that group. Upon receiving a reward, the posterior distributions for the dataset and the dataset group are updated to $\mathcal{N}(\mu', \sigma'^2)$ and $\mathcal{N}(\theta', \hat{\sigma}'^2)$ respectively. After training, dataset groups and datasets with higher posterior means are selected as described in Section \ref{dataset selection}. }
    \label{fig:diagram}
\end{figure*}

\subsection{\modelname{} Initialization}
To address this selection objective, we introduce \modelname{}, a bi-level hierarchical Bayesian model that captures structured uncertainty across data groups and individual datasets. As depicted in Figure \ref{fig:diagram}, each data group $g_i$ is modeled with a latent parameter $\theta_i$ encoding its expected utility, and each dataset $d_{i,j}$ is governed by a local parameter $\theta_{i,j}$, with corresponding reward observations $r_{i,j}(t)$ at timestep $t$. We assume normal distributions for both the priors and the reward models, with unknown means and fixed variances. Conditional on $\theta_{i,j}$, the reward $r_{i,j}(t)$ is independent of the group-level parameter $\theta_i$. The generative process is:
\begin{equation}\label{eq:1}
\begin{split}
\theta _ {i} & \sim   \mathcal{N}(\mu_i , \sigma_i^2),  \forall i\in [n] \\
\theta _ {i,j} | \theta_{i} & \sim  \mathcal{N}(\theta _ {i}, \hat{\sigma}_i^2), \forall j\in [m] \\
r_{i,j}(t) | \theta _ {i,j} & \sim  \mathcal{N}(\theta_{i,j} , \sigma_r^2), \forall D(t)\!=\!d_{i,j},
\end{split}
\end{equation} 
where $\mu_i$ is the mean of the prior distribution for data group $g_i$, $\sigma_i^2$  is the variance of the group prior, $\hat{\sigma}_i^2$ is the variance of the dataset prior $\theta_{i,j}$, and $\sigma_r^2$ is the variance of the reward observation model.
The goal is to iteratively update the posterior distribution of $\theta_i$ and $\theta_{i,j}$ by incorporating all observed reward values accumulated up to the current time step $t$. Through this continual update process, \modelname{} converges towards accurate estimations of the true distributions for both $\theta_i$ and $\theta_{i,j}$ after a number of iterations, as described in Algorithm \ref{HB} in the Appendix.
Initialization begins with all dataset groups $\mathbf{g}$ sharing a common prior $\mathcal{N}(\mu_0 , \sigma_0^2)$ and $\mathcal{N}(\theta_{0}, \hat{\sigma}_0^2)$. At each time step $t$, $\hat{\theta}_{i}$ is drawn from the normal distributions associated with each dataset group $\hat{\theta}_{i}  \sim  P(\theta_i|r_i)$ and the dataset group $g_i$ with the largest value is chosen. Given dataset group selection $g_i$, \modelname{} then draws $\hat{\theta}_{i,j}$ from the distributions associated with the datasets within the chosen dataset group, \ie $\hat{\theta} _ {i,j}  \sim  P(\theta_{i,j}|r_{i,j})$, and selects the dataset with the largest values, denoted as $D(t)\!=\!d_{i,j}$. 

\subsection{\modelname{} Posterior Computation}
\modelname{} receives a reward from the chosen dataset and updates the distribution associated with the chosen dataset group and dataset using Eqs. (\ref{eq:3}) and (\ref{eq:4}). 
The posterior distribution of $\theta_i$ after observing reward values $r_i=\{r_{i,j}\}, j \in [m]$, where $r_{i,j}\!=\!\{r_{i,j}(t), \forall D(t)\!=\!d_{i,j}\}$, is given by:
\begin{equation} \label{eq:2}
\scalemath{0.85}{
\int_{\theta_{i,j}}\left( \prod_{j=1}^m  \mathcal{N}(r_ {i,j} ;  \theta_{i,j} ,  \sigma_r ^ {2}) \right)  \mathcal{N}(  \theta_{i,j}  ; \theta _ {i}, \hat{\sigma}_i^2 )d  \theta_{i,j}   \mathcal{N}( \theta _ {i}  ;\mu_i , \sigma_i^2).
}
\end{equation}

\noindent From Eq.(\ref{eq:2}), this yields the closed-form posterior:
\begin{equation} \label{eq:3}
P(\theta_i |r_{i})\!=\!\mathcal{N}\left(\lambda_i^2 \left(\frac{\mu_i}{\sigma_i^2}+ \frac{\bar{s_i}}{ \hat{\sigma_i}^2 + \frac{\sigma_r ^ {2}}{n_i}}\right),\lambda_i^2\right)
\end{equation} 
where 
\begin{equation}
\lambda_i^2 \!=\!\left(\frac{1}{\sigma_i^2} +  \frac{1}{\hat{\sigma_i}^2+ \frac{\sigma_r ^ {2}}{n_i}}\right)^{-1}, \quad \bar s_i\!=\!\frac{\sum_{j=1}^{m}r_{i,j}}{n_i}.
\end{equation}
Here, $n_i$ is the total number of selections for group $g_i$, and $\bar{s}_i$ is the aggregated mean reward across datasets in group $i$. The posterior mean is a precision-weighted average of the prior mean $\mu_i$ and the empirical group mean $\bar{s}_i$. The influence of the prior decays with more observations as $\lambda_i^2$ decreases.
Since the reward $r_{i,j}(t)$ is conditionally independent of the data group parameter $\theta_i$, the posterior density of $\theta_{i,j}$, after observing rewards $r_{i,j}(t)$ at time step $t$, is computed by:
\begin{equation}
P(\theta_{i,j} \mid r_{i,j}) \propto P(\theta_{i,j}) \prod_{t: D(t) = d_{i,j}}\!\mathcal{N}(r_{i,j}(t); \theta_{i,j}, \sigma_r^2),
\end{equation}
resulting in the posterior:
\begin{equation} \label{eq:4}
P(\theta_{i,j} \mid r_{i,j})\!=\!\mathcal{N}\left(
\lambda_{i,j}^2 \left( \frac{\theta_i}{\hat{\sigma}_i^2} + \frac{\bar{s}_{i,j} \cdot n_{i,j}}{\sigma_r^2} \right),
\lambda_{i,j}^2
\right)
\end{equation}
where 
\begin{equation}
    \lambda_{i,j}^2 \!=\!\left(\frac{1}{\hat {\sigma }_i^ {2}} +  \frac{n_{i,j}}{ \sigma_r ^ {2}}\right)^{-1}, \quad \bar s_{i,j} \!=\!\frac{r_{i,j}}{n_{i,j}}
\end{equation}
Here, $n_{i,j}$ is the number of times dataset $d_{i,j}$ has been selected and  $\bar{s}_{i,j}$ empirical mean of $r_{i,j}$.

Different from the dataset group posterior, the dataset posterior only depends on the rewards received by the dataset. 
Similar to the dataset group prior mean $\mu_i$, $\theta_i$ is a bias term that influences the decay of the dataset posterior mean. As $n_{i,j} \to \infty$, the dataset posterior variance goes to zero, and the dataset posterior mean approaches $\bar s_{i,j}$.

\subsection{Dataset Selection Based on Posterior Distributions} \label{dataset selection}

We formalize dataset selection using posterior means in a two-step process: first selecting a dataset group, then a dataset within that group. A dataset or group is selected if its posterior mean $\mu$ exceeds a percentile-based threshold, \ie if $\mu > F^{-1}(x)$, where $F^{-1}$ is the inverse cumulative distribution function (CDF) over the posterior means, setting the threshold at the $x$-th percentile. The selection threshold \( x \) is adaptively chosen based on the specific needs and constraints of the training environment. For example, a high percentile (\eg 90th) indicates a stringent criterion, suitable for scenarios with high training costs or where poor data quality significantly impacts model performance. Conversely, a lower percentile may be used in exploratory settings or when additional data inclusion costs are minimal.
Alternatively, based on the use case, the selection of top-$x$ datasets or dataset groups may be more appropriate.

\subsection{Algorithmic Complexity}
At each selection step, \modelname{} performs two sequential operations: 
(1) inter-group sampling by drawing $\hat{\theta}_i\!\sim\!P(\theta_i \mid r_i)$ 
for all $n$ groups, and 
(2) intra-group sampling by drawing 
$\hat{\theta}_{i,j}\!\sim\!P(\theta_{i,j} \mid r_{i,j})$ 
for the $m_i$ datasets in the chosen group. 
This yields a per-step computational cost of $O(n + m_i)$. 
Posterior updates for the chosen dataset and group require constant time per step, 
as the closed-form updates in Eqs.~(4) and (7) avoid iterative optimization.\looseness-1

By contrast, a flat selection strategy must evaluate all 
$|D|\!=\!\sum_{i=1}^n m_i$ datasets at each step, incurring $O(|D|)$ cost. 
When groups are large, the hierarchical formulation amortizes exploration: 
feedback from a single dataset selection updates both its dataset-level 
and group-level posteriors, effectively sharing information across datasets 
in the same group. This reduces the total number of dataset evaluations 
required to achieve a fixed target accuracy, as consistently demonstrated in our experiments.\looseness-1

\section{Experiments} \label{sec:experiments}

\noindent \textbf{Datasets.}
We validate \modelname{} on two widely used benchmarks in domain adaptation: \textbf{\textsc{Digit-Five}} and \textbf{\textsc{DomainNet}}~\cite{peng2019moment}. Each dataset contains multiple domain-specific subsets for a shared classification task. \textsc{Digit-Five} includes digit images from five domains (\textsc{MNIST}, \textsc{MNIST-M}, \textsc{USPS}, \textsc{SVHN}, and \textsc{SYN}), while \textsc{DomainNet} comprises object recognition images across different styles (\textsc{CLIPART}, \textsc{QUICKDRAW}, \textsc{REAL}, and \textsc{SKETCH}). Each domain is divided into three disjoint subsets to simulate distributed or federated settings. We use preprocessed versions of these datasets from \citet{schrod2023fact}, where fixed-size feature vectors are extracted from images for training and evaluation.

To evaluate the robustness of \modelname{} across varying dataset compositions, we examine two grouping strategies. In the \textbf{perfect group} setting, each group contains three subsets from the same domain (\eg {mn0, mn1, mn2} from \textsc{MNIST}),  modeling cases where repositories or institutions curate domain-specific datasets. In the \textbf{mixed group} setting, subsets from different domains are combined into groups (\eg {mn1, mn2, mm0}),  modeling cases where datasets from multiple sources or domains are aggregated for a shared task and group assignments are noisy or imperfect. Preprocessing steps, group definitions, and dataset statistics are provided in the Appendix.
\begin{figure}[t!]
    \begin{subfigure}[t]{0.43\textwidth}
        \raggedleft
        \includegraphics[width=\linewidth]{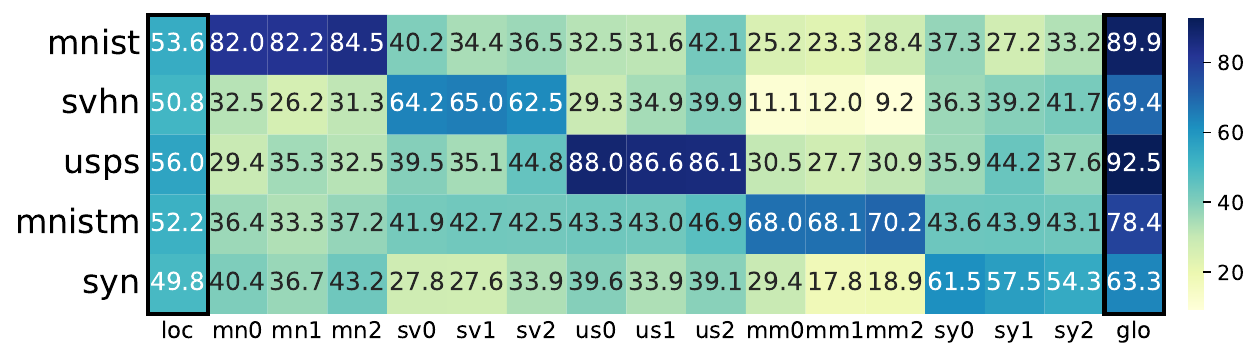}
        \caption{\textsc{Digit-Five}}
        \vspace{0.3cm}
    \end{subfigure}
    \begin{subfigure}[t]{0.45\textwidth}
        \raggedright
        \includegraphics[width=\linewidth]{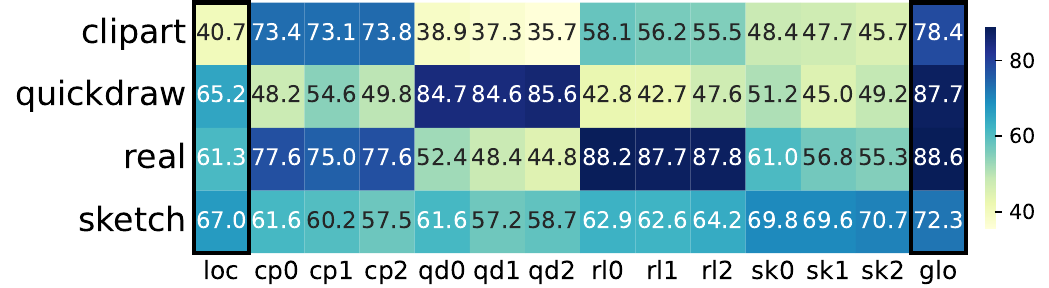}
        \caption{\textsc{DomainNet}}
    \end{subfigure}
    \caption{Accuracy heatmaps of local classifiers after training on different \textsc{Digit-Five} and \textsc{DomainNet} subsets.  The first column shows local test accuracy for each subset. The last column indicates the optimal accuracy achievable when training on all available relevant same-domain datasets. Middle columns depict accuracy after augmenting training data with additional subsets from same and different domains.}
    \label{fig:gt_acc}
\end{figure}

\paragraph{Implementation Details.} For \textsc{Digit-Five}, each local model is a lightweight CNN trained on its respective domain-specific subsets (\eg \textsc{MNIST}, \textsc{SVHN}), while for \textsc{DomainNet}, local models are three-layer multilayer perceptrons (MLPs). Local accuracy refers to model performance on its own domain without any additional training. Additional implementation details are provided in the Appendix.

Figure~\ref{fig:gt_acc} summarizes the empirical results obtained by training local models on different external datasets. These ground-truth results serve as a reference for evaluating the potential benefit of dataset selection. In \textsc{Digit-Five}, models trained on external datasets consistently underperform compared to their local baselines, indicating strong domain-specific bias. In contrast, \textsc{DomainNet} exhibits more favorable cross-domain transfer; for example, training the \textsc{REAL} classifier on subsets from \textsc{CLIPART} yields noticeable performance gains. This distinction underscores the practical relevance of dataset selection in heterogeneous sharing scenarios.

\begin{table*}[t!] 
\centering
\resizebox{\linewidth}{!}{
\begin{tabular}{@{}lclllllll}
\toprule
\textbf{Method} & \textbf{Hierarchical} & \textbf{\textsc{MNIST}} & \textbf{\textsc{SVHN}} & \textbf{\textsc{USPS}} & \textbf{\textsc{MNIST-M}} & \textbf{\textsc{SYN}} & \textbf{AVG} \\ \midrule
Local & {\color{red} \ding{55}} & 52.7$\pm$6.5& 50.9$\pm$3.4 & 52.2$\pm$3.2 & 49.4$\pm$2.4 & 50.9$\pm$5.1 & 51.2$\pm$4.1\\
Global & {\color{red} \ding{55}} & 89.3$\pm$1.1& 69.7$\pm$1.4 & 92.2$\pm$0.7 & 80.2$\pm$1.1 & 62.8$\pm$2.8 & 78.8$\pm$1.4\\ \midrule
Core-sets~\cite{sener2018active} & {\color{red} \ding{55}} & 75.7$\pm$2.3  \textcolor{red}{$\downarrow$13.8} & 52.8$\pm$2.7 \textcolor{red}{$\downarrow$16.4}&74.0$\pm$3.6 \textcolor{red}{$\downarrow$17.2}&60.8$\pm$2.1 \textcolor{red}{$\downarrow$18.1}&40.8$\pm$1.9 \textcolor{red}{$\downarrow$22.1}&60.8$\pm$2.5 \textcolor{red}{$\downarrow$17.5}\\
FreeSel~\cite{xie2023towards} & {\color{red} \ding{55}} & 87.6$\pm$1.2 \textcolor{red}{$\downarrow$1.9}& 39.3$\pm$4.0 \textcolor{red}{$\downarrow$29.9}&29.3$\pm$3.1 \textcolor{red}{$\downarrow$61.9}&65.4$\pm$2.2 \textcolor{red}{$\downarrow$13.5}&40.7$\pm$2.9 \textcolor{red}{$\downarrow$22.2}&52.5$\pm$2.7 \textcolor{red}{$\downarrow$25.8}\\
ActiveFT~\cite{xie2023active} & {\color{red} \ding{55}} & 58.2$\pm$1.6\textcolor{red}{$\downarrow$31.3} & 
53.6$\pm$1.6\textcolor{red}{$\downarrow$15.6} & 59.2$\pm$1.3\textcolor{red}{$\downarrow$32.0} & 48.3$\pm$0.9\textcolor{red}{$\downarrow$30.6} & 41.4$\pm$1.5\textcolor{red}{$\downarrow$21.5} & 52.1$\pm$1.4\textcolor{red}{$\downarrow$26.2} \\
BiLAF~\cite{luboundary} &{\color{red} \ding{55}}& 62.6$\pm$0.5\textcolor{red}{$\downarrow$26.9} & 56.8$\pm$0.4\textcolor{red}{$\downarrow$12.4} & 67.3$\pm$0.5\textcolor{red}{$\downarrow$23.9} & 50.1$\pm$0.5\textcolor{red}{$\downarrow$28.8} & 52.6$\pm$1.0\textcolor{red}{$\downarrow$10.3} & 57.9$\pm$0.6\textcolor{red}{$\downarrow$20.4} \\
\hline
\rowcolor{lightgray!30} \textbf{\modelname{}} & {\color{ForestGreen}\textbf{\ding{51}}} & \textbf{89.5$\pm$0.6} & \textbf{69.2$\pm$3.4} & \textbf{91.2$\pm$0.9} & \textbf{78.9$\pm$0.5} & \textbf{62.9$\pm$1.6}  & {\textbf{78.3$\pm$1.4}}  \\

\bottomrule
\end{tabular}
}
\caption{Performance comparison on \textsc{Digit-Five} against baselines (averaged over 5 runs) Best performance is \textbf{{bold}}. Red downward arrows (\textcolor{red}{$\downarrow$}) indicate absolute drops in accuracy relative to the best-performing method. 
}\label{tab:digit_five_accuracy}
\end{table*}

\begin{table*}[t!] 
\centering
\resizebox{\linewidth}{!}{
\begin{tabular}{@{}lclllll}
\toprule
\textbf{Method} & \textbf{Hierarchical} & \textbf{\textsc{CLIPART}} & \textbf{\textsc{QUICKDRAW}} & \textbf{\textsc{REAL}} & \textbf{\textsc{SKETCH}} & \textbf{AVG} \\ \midrule
Local &  {\color{red} \ding{55}}& 40.0$\pm$2.4& 64.0$\pm$2.1& 61.0$\pm$1.1 & 67.5$\pm$1.1  & 58.1$\pm$1.7\\
Global &  {\color{red} \ding{55}} & 78.5$\pm$0.6& 86.7$\pm$0.5& 88.4$\pm$0.6 & 72.3$\pm$0.8  & 81.6$\pm$1.1\\ \midrule

Core-sets~\cite{sener2018active} & {\color{red} \ding{55}} & 59.1$\pm$0.9 \textcolor{red}{$\downarrow$18.2} & 74.1$\pm$0.3 \textcolor{red}{$\downarrow$12.3}&80.1$\pm$0.6 \textcolor{red}{$\downarrow$8.3}&67.6$\pm$0.4 \textcolor{red}{$\downarrow$4.2}&70.2$\pm$0.6 \textcolor{red}{$\downarrow$10.8}\\
FreeSel~\cite{xie2023towards} & {\color{red} \ding{55}} & 70.1$\pm$2.1 \textcolor{red}{$\downarrow$7.2} & 81.7$\pm$0.8 \textcolor{red}{$\downarrow$4.6}&85.6$\pm$0.7 \textcolor{red}{$\downarrow$2.8}&67.2$\pm$1.3 \textcolor{red}{$\downarrow$4.6}&77.7$\pm$1.2 \textcolor{red}{$\downarrow$3.3}\\
ActiveFT~\cite{xie2023active} & {\color{red} \ding{55}} & 67.6$\pm$1.8\textcolor{red}{$\downarrow$9.7} & 78.0$\pm$1.0\textcolor{red}{$\downarrow$8.3} & 83.8$\pm$1.1\textcolor{red}{$\downarrow$4.6} & 67.8$\pm$1.1\textcolor{red}{$\downarrow$4.0} & 74.3$\pm$1.3\textcolor{red}{$\downarrow$6.7} \\
BiLAF~\cite{luboundary} &{\color{red} \ding{55}} & 69.0$\pm$1.6\textcolor{red}{$\downarrow$8.3} & 81.3$\pm$0.5\textcolor{red}{$\downarrow$5.0} & 85.8$\pm$0.5\textcolor{red}{$\downarrow$2.6} & 67.8$\pm$0.7\textcolor{red}{$\downarrow$4.0} & 76.0$\pm$0.8\textcolor{red}{$\downarrow$5.0} \\
\hline
\rowcolor{lightgray!30} \textbf{\modelname{}} & {\color{ForestGreen}\textbf{\ding{51}}} & \textbf{77.3$\pm$0.8} &\textbf{86.3$\pm$1.1}& \textbf{88.4$\pm$0.8} & \textbf{71.8$\pm$0.9} & \textbf{81.0$\pm$0.9} \\ 

\bottomrule
\end{tabular}
}
\caption{Performance comparison on \textsc{DomainNet} against baselines (averaged over 5 runs). Best performance is \textbf{{bold}}.  Red downward arrows (\textcolor{red}{$\downarrow$}) indicate absolute drops in accuracy relative to the best-performing method. 
} \label{tab:domain_accuracy_stop}
\end{table*}

\paragraph{Baselines.}
We compare against existing methods to assess: (1) \modelname{}’s effectiveness in dataset selection relative to state-of-the-art data selection approaches, and (2) its ability to capture dependencies among datasets. 

\noindent \textbf{Core-sets} \cite{sener2018active}, which selects representative samples via geometric coverage, such that models learned only on the selected subset are as competitive. 

\noindent \textbf{FreeSel} \cite{xie2023towards}, uses a pretrained vision transformer to perform one‐pass, supervision‐free data selection,  with a time efficiency close to random selection.

\noindent \textbf{ActiveFT} \cite{xie2023active}, which optimizes selection to match the data distribution while preserving diversity.

\noindent \textbf{BiLAF} \cite{luboundary}, extends ActiveFT by introducing boundary uncertainty to enable one-shot label-free selection through pseudo-class estimation and iterative refinement. 

\noindent In addition, we include two baselines for reference: \textbf{Local}, trained only on local data, and \textbf{Global}, trained on all datasets from the same domain, representing lower and upper bounds.

\begin{figure*}[t]
\centering
\begin{tikzpicture}
\begin{axis}[
  width=0.48\linewidth, height=5.4cm,
  xlabel={Steps (more $\rightarrow$ fewer)}, ylabel={Accuracy (\%)},
  xmin=120, xmax=230, x dir=reverse,
  ymin=55, ymax=95,
  xmajorgrids, ymajorgrids, grid style=dashed, tick align=inside,
  title={\textsc{Digit-Five}},
    title style={
  at={(axis description cs:0.5,-0.3)},
  anchor=north,
},
    legend columns=5,
    legend image post style={draw=gray!70, fill=gray!70},
  legend style={
    at={(0.5,1)}, anchor=south, draw=none,
    font=\footnotesize, column sep=0.4em
  },
]
\addplot[only marks, color=orange!60, mark=*,  mark size=2.3pt] coordinates {(166,88.5)}; 
\addplot[only marks, color=orange!60, mark=square*,  mark size=2.3pt] coordinates {(224,69.9)}; 
\addplot[only marks, color=orange!60, mark=triangle*,mark size=2.5pt] coordinates {(163,90.9)}; 
\addplot[only marks, color=orange!60, mark=diamond*, mark size=2.5pt] coordinates {(188,77.6)}; 
\addplot[only marks, color=orange!60, mark=star,     mark size=2.3pt] coordinates {(208,58.7)}; 
\addplot[only marks, color=blue!70,  mark=*,        mark size=2.3pt] coordinates {(137,89.5)}; 
\addplot[only marks, color=blue!70,  mark=square*,  mark size=2.3pt] coordinates {(183,69.2)}; 
\addplot[only marks, color=blue!70,  mark=triangle*,mark size=2.5pt] coordinates {(140,91.2)}; 
\addplot[only marks, color=blue!70,  mark=diamond*, mark size=2.5pt] coordinates {(154,78.9)}; 
\addplot[only marks, color=blue!70,  mark=star,     mark size=2.3pt] coordinates {(194,62.9)}; 
\addplot[only marks, color=teal!70,  mark=*,        mark size=2.5pt] coordinates {(133,88.5)}; 
\addplot[only marks, color=teal!70,  mark=square*,  mark size=2.5pt] coordinates {(173,69.8)}; 
\addplot[only marks, color=teal!70,  mark=triangle*,mark size=2.7pt] coordinates {(143,90.4)}; 
\addplot[only marks, color=teal!70,  mark=diamond*, mark size=2.7pt] coordinates {(165,77.8)}; 
\addplot[only marks, color=teal!70,  mark=star,     mark size=2.5pt] coordinates {(187,58.6)}; 

\addlegendimage{only marks, mark=*, color=black}  \addlegendentry{MNIST}
\addlegendimage{only marks, mark=square*}   \addlegendentry{SVHN}
\addlegendimage{only marks, mark=triangle*} \addlegendentry{USPS}
\addlegendimage{only marks, mark=diamond*}  \addlegendentry{MNIST-M}
\addlegendimage{only marks, mark=star}      \addlegendentry{SYN}

\end{axis}
\end{tikzpicture}
\hfill
\begin{tikzpicture}
\begin{axis}[
  width=0.48\linewidth, height=5.4cm,
  xlabel={Steps (more $\rightarrow$ fewer)}, ylabel={Accuracy (\%)},
  xmin=140, xmax=220, x dir=reverse,
  ymin=68, ymax=90,
  xmajorgrids, ymajorgrids, grid style=dashed, tick align=inside,
  title={\textsc{DomainNet}},
  title style={
  at={(axis description cs:0.5,-0.3)},
  anchor=north,
},
    legend columns=4,
    legend image post style={draw=gray!70, fill=gray!70},
  legend style={
    at={(0.5,1)}, anchor=south, draw=none,
    font=\footnotesize, column sep=0.4em
  },
]
\addplot[only marks, color=orange!60, mark=*,        mark size=2.3pt] coordinates {(215,77.5)}; 
\addplot[only marks, color=orange!60, mark=square*,  mark size=2.3pt] coordinates {(180,87.0)}; 
\addplot[only marks, color=orange!60, mark=triangle*,mark size=2.5pt] coordinates {(175,88.3)}; 
\addplot[only marks, color=orange!60, mark=diamond*, mark size=2.5pt] coordinates {(177,70.7)}; 
\addplot[only marks, color=blue!70,  mark=*,        mark size=2.3pt] coordinates {(184,77.3)}; 
\addplot[only marks, color=blue!70,  mark=square*,  mark size=2.3pt] coordinates {(166,86.3)}; 
\addplot[only marks, color=blue!70,  mark=triangle*,mark size=2.5pt] coordinates {(164,88.4)}; 
\addplot[only marks, color=blue!70,  mark=diamond*, mark size=2.5pt] coordinates {(168,71.8)}; 
\addplot[only marks, color=teal!70,  mark=*,        mark size=2.5pt] coordinates {(186,77.3)}; 
\addplot[only marks, color=teal!70,  mark=square*,  mark size=2.5pt] coordinates {(150,86.9)}; 
\addplot[only marks, color=teal!70,  mark=triangle*,mark size=2.7pt] coordinates {(190,88.8)}; 
\addplot[only marks, color=teal!70,  mark=diamond*, mark size=2.7pt] coordinates {(158,70.5)}; 

\addlegendimage{only marks, mark=*}         \addlegendentry{CLIPART}
\addlegendimage{only marks, mark=square*}   \addlegendentry{QUICKDRAW}
\addlegendimage{only marks, mark=triangle*} \addlegendentry{REAL}
\addlegendimage{only marks, mark=diamond*}  \addlegendentry{SKETCH}
\end{axis}
\end{tikzpicture}
\caption{Pareto trade-offs between accuracy and selection cost. Each point is a method–domain result (\textsc{Digit-Five} left, \textsc{DomainNet} right). Marker shape encodes the domain, while color distinguishes the methods: \textbf{\textcolor{orange!60}{\dasflat{}}}, \textbf{\textcolor{teal!70}{\modelname{} (mixed)}}, and \textbf{\textcolor{blue!70}{\modelname{}}}. Points toward the upper-right represent better trade-offs (higher accuracy, fewer steps). Across both benchmarks, the upper-right region is occupied by hierarchical variants with  \textbf{\textcolor{blue!70}{\modelname{}} }contributing most of the frontier on \textsc{Digit-Five} and sharing the frontier with \textbf{\textcolor{teal!70}{\modelname{} (mixed)}} on \textsc{DomainNet}.}
\label{fig:pareto}
\end{figure*}

\subsection{Experimental Results}\label{results}
\noindent  Table~\ref{tab:digit_five_accuracy} reports mean and standard deviation over five independent runs on \textsc{Digit-Five} subdomains, where we compare \modelname{} to local and global baselines as well as the four state-of-the-art data selection baselines.
Across all five domains, \modelname{} matches the global model, achieving an average accuracy of 78.3\%, which is only 0.5\% below the global upper bound (78.8\%) and significantly higher than the local lower bound (51.2\%). These results indicate that our method is capable of effectively leveraging heterogeneous data sources.

Compared to competitive baselines, \modelname{} exhibits substantial gains. For instance, FreeSel underperforms by over 25.8\% on average, and notably degrades performance on \textsc{SVHN}, \textsc{USPS}, and \textsc{SYN}, suggesting that its model-free selection policy does not work well under our problem setting where the selection pool contains irrelevant data. Similarly, ActiveFT and BiLAF fall behind by 26.2\% and 20.4\%, respectively. Notably, these methods exhibit particularly low accuracy on \textsc{MNIST-M} and \textsc{SYN}, which represent domains with significant distributional divergence from the rest of the datasets. This performance drop suggests that baselines struggle to generalize when the target domain is poorly aligned with the source distribution, highlighting their limitations in handling high domain shift scenarios.
In contrast, \modelname{} consistently maintains top performance with low variance, highlighting its robustness across target domains.

Table~\ref{tab:domain_accuracy_stop} shows results on \textsc{DomainNet}. While performance margins are narrower than in \textsc{Digit-Five}, \modelname{} still outperforms all baselines by 3.3--10.8\%. This is likely because all models use features extracted from a ResNet-18 backbone that was pretrained on the combined dataset. The shared feature extractor reduces the distributional differences between domains, making the task inherently easier for all methods and diminishing relative gains. Nevertheless, \modelname{} maintains its advantage, underscoring its effectiveness even when inter-domain variation is minimized.

\section{Ablation Studies}\label{sec:ablations}
To better understand the contributions of individual components in \modelname{} and the conditions under which it is most effective, we conduct a series of ablation studies. These experiments are designed to (1) isolate the effect of hierarchical modeling, (2) assess robustness to imperfect group definitions, (3) evaluate the role of Bayesian posterior updates, and (4) examine sensitivity to the exploration–exploitation trade-off. We also examine (5) the impact of selection granularity and (6) quantify efficiency gains from each design choice.

\subsection{Impact of Hierarchical Grouping} 
To understand the importance of hierarchical grouping, we compare \modelname{} against two baseline variants: \dasflat{}, a non-hierarchical counterpart, and \modelname{} (mixed), which uses imperfect group assignments. Figure~\ref{fig:pareto} presents Pareto frontiers of accuracy versus selection cost (exploration steps) for each domain in \textsc{Digit-Five} and \textsc{DomainNet}, with marker shapes indicating domains and colors indicating methods.
Compared to the non-hierarchical \dasflat{}, \modelname{} consistently delivers equal or higher accuracy at substantially lower selection cost. On \textsc{Digit-Five}, this translates to savings of 20–60 steps per domain without sacrificing accuracy. 
When compared to \modelname{} (mixed), the gap is small in most domains, with the mixed variant often lying on or near the Pareto frontier achieved by perfect grouping. This indicates that \modelname{} is robust to imperfect group assignments, with only modest performance drops in more challenging domains like \textsc{SYN}, \textsc{QUICKDRAW}, and \textsc{REAL}. Overall, these results show that hierarchical grouping not only improves efficiency and accuracy but also maintains strong performance under noisy or partially incorrect group structures.

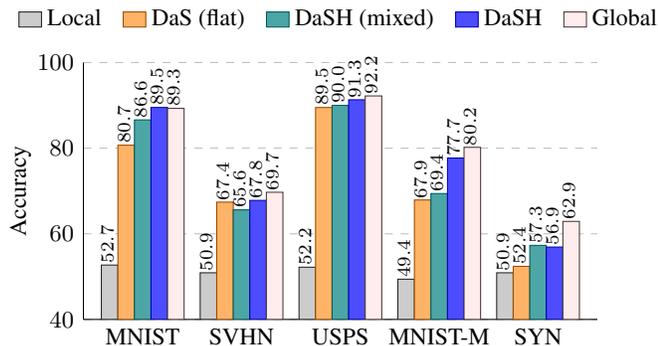
\begin{figure}[t]
\centering
\begin{tikzpicture}
    \begin{axis}[
        ybar=0pt,
        ymin=40,
        ymax=100,
        width=\columnwidth,
        height=5cm,
        enlarge x limits=0.15,
        ymajorgrids=true,
        grid style=dashed,
        symbolic x coords={MNIST, SVHN, USPS, MNIST-M, SYN},
        xtick=data,
        xticklabel style={yshift=1ex, font=\footnotesize},
        yticklabel style={font=\footnotesize},
        tick style={draw=none},
        ylabel={Accuracy},
        ylabel near ticks,
        ylabel style={font=\footnotesize, yshift=-1ex},
        axis x line*=bottom,
        axis y line*=left,
        bar width=0.22cm,
        nodes near coords,
        every node near coord/.append style={
            font=\scriptsize,
            rotate=90,
            xshift=8pt,
            yshift=-5pt,
            color=black,
            /pgf/number format/.cd,
                fixed,
                fixed zerofill,
                precision=1
        },
        legend style={
        at={(0.5,1.25)},
        anchor=north,
        legend columns=-1,
        draw=none,
        column sep=0.001em,
        font=\footnotesize,
        /tikz/every even column/.append style={column sep=6pt} 
    },
        legend image code/.code={
            \draw[#1] (0cm,-0.1cm) rectangle (0.3cm,0.1cm);
        },
    ]

    \addplot[gray!30!black,fill=gray!40!white] coordinates {(MNIST, 52.7) (SVHN,50.9) (USPS,52.2) (MNIST-M,49.4) (SYN,50.9)};
    \addlegendentry{Local}

    \addplot[orange!60!black,fill=orange!60!white] coordinates {(MNIST, 80.7) (SVHN,67.4) (USPS,89.5) (MNIST-M,67.9) (SYN,52.4)};
    \addlegendentry{\dasflat{}}

    \addplot[teal!70!black,fill=teal!70!white] coordinates {(MNIST, 86.6) (SVHN,65.6) (USPS,90.0) (MNIST-M,69.4) (SYN,57.3)};
    \addlegendentry{\modelname{} (mixed)}

    \addplot[blue!70!black,fill=blue!70!white] coordinates {(MNIST, 89.5) (SVHN,67.8) (USPS,91.3) (MNIST-M,77.7) (SYN,56.9)};
    \addlegendentry{\modelname{}}

    \addplot[pink!30!black,fill=pink!30!white] coordinates {(MNIST, 89.3) (SVHN,69.7) (USPS,92.2) (MNIST-M,80.2) (SYN,62.9)};
    \addlegendentry{Global}

    \end{axis}
\end{tikzpicture}
\caption{Performance under budget constraints. Under limited exploration (15 steps), \modelname{} and \modelname{} (mixed) outperform \dasflat{} on 4 out of 5 datasets. \textit{Local} and \textit{Global} denote the lower and upper bounds, respectively.}
\label{fig:steps_15}
\end{figure}

\subsection{Comparison Under Limited Exploration}
We evaluate the ability of each method to identify useful datasets under stringent exploration budgets. Specifically, each method explores each dataset only once, totaling 15 steps across the 15 datasets in \textsc{Digit-Five}. Figure~\ref{fig:steps_15} reports the resulting accuracy for each domain. Under this extreme budget constraint, both \modelname{} and \modelname{} (mixed) outperform the non-hierarchical \dasflat{} in 4 out of 5 domains. The gains over \dasflat{} are substantial:
+8.8\% on \textsc{MNIST}, +1.8\% on \textsc{USPS}, +9.8\% on \textsc{MNIST-M}, and +4.5\% on \textsc{SYN}.
Even with imperfect grouping, \modelname{} (mixed) closely tracks the performance of perfect grouping, with accuracy differences within 1–2\% in most domains. The {Local} and {Global} baselines show that hierarchical variants close more than half the gap to the global optimum despite operating under a 15-step budget. These results confirm that hierarchical grouping enables efficient, high-quality dataset selection even under severe exploration limits.\looseness-1

\subsection{Effectiveness Under Weak Initialization}
We additionally investigate whether \modelname{} can enhance performance when initial local model accuracy is very low. We train initial local classifiers using 10\%, 20\%, and 50\% of the available training data. Table~\ref{tab:low_local_acc} shows consistent accuracy gains across all conditions, even when initial accuracy is as low as 9.6\% (\textsc{USPS}), demonstrating \modelname{}'s robustness to significant variations in initial performance before selection.
\begin{table}[t]
\centering
\resizebox{\columnwidth}{!}{
\begin{tabular}{lcccccc}
\toprule
\textbf{\% Train} & \multicolumn{2}{c}{\textbf{10\%}} & \multicolumn{2}{c}{\textbf{20\%}} & \multicolumn{2}{c}{\textbf{50\%}} \\
\cmidrule(r){2-3} \cmidrule(r){4-5} \cmidrule(r){6-7}
\textbf{Name}& \textbf{Init.} & \textbf{\modelname{}} &\textbf{ Init.} & \textbf{\modelname{}} & \textbf{Init.} & \textbf{\modelname{}} \\
\midrule
\textsc{MNIST} & 17.6 & 31.5 & 23.6 & 89.6 & 36.6 & 89.6 \\
\textsc{SVHN} & 12.8 & 24.2 & 21.2 & 21.5 & 35.6 & 66.7 \\
\textsc{USPS} & 9.6  & 13.5 & 12.8 & 28.6 & 31.2 & 91.4 \\
\textsc{MNIST-M} & 20.6 & 55.1 & 28.8 & 57.6 & 44.2 & 79.3 \\
\textsc{SYN} & 26.6 & 37.6 & 21.4 & 24.9 & 27.4 & 41.0 \\
\bottomrule
\end{tabular}
}
\caption{\modelname{} improves performance even with a weak initial model with low accuracy. This table reports accuracy on \textsc{Digit-Five} when initially trained on 10\%, 20\%, and 50\% of the local training data (\textbf{Init.}), and after using \modelname{} to select additional datasets for training (\textbf{\modelname{}}).} 
\label{tab:low_local_acc}
\end{table}

\begin{figure*}[t!]
    \centering
     \resizebox{0.9\linewidth}{!}{%
    
    \begin{tabular}{@{}c*{5}{@{\hskip 0.5em}c}@{\hskip 1.5em}*{5}{@{\hskip 0.5em}c}@{}}
        \textbf{ Methods} & \multicolumn{5}{c}{\textbf{Digit-Five}} 
        & \multicolumn{5}{c}{\textbf{DomainNet}} \\[0.3em]

        ~ & \textsc{SYN} & \textsc{SVHN} & \textsc{SVHN} & \textsc{MNIST-M} & \textsc{USPS}
          & \makebox[1.5cm][c]{\scriptsize \textsc{QUICKDRAW}} & \makebox[1.5cm][c]{\scriptsize \textsc{QUICKDRAW}} & \makebox[1.5cm][c]{\scriptsize \textsc{REAL}} & \makebox[1.5cm][c]{\scriptsize \textsc{REAL}} & \makebox[1.5cm][c]{\scriptsize \textsc{SKETCH}} \\[-0.2em]
        \raisebox{0.4cm}{\textbf{Core-Sets}}
        & \redimg{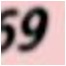}
        & \redimg{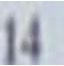}
        & \redimg{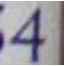}
        & \redimg{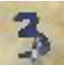}
        & \redimg{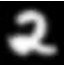}
        & \redimg{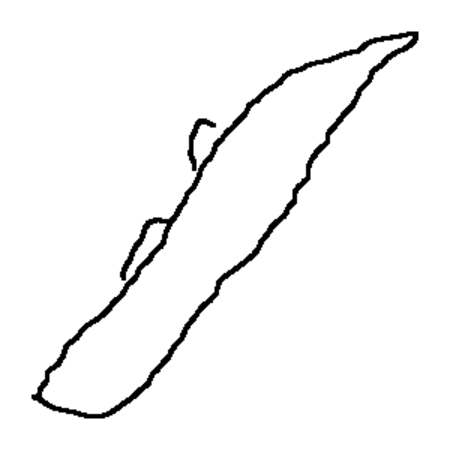}
        & \redimg{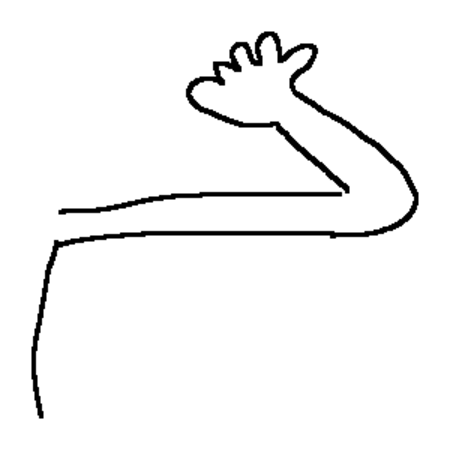}
        & \redimg{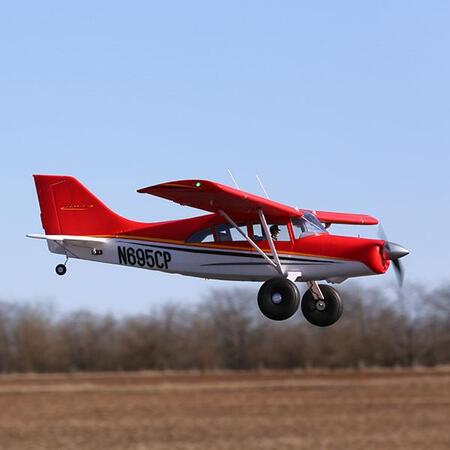}
        & \redimg{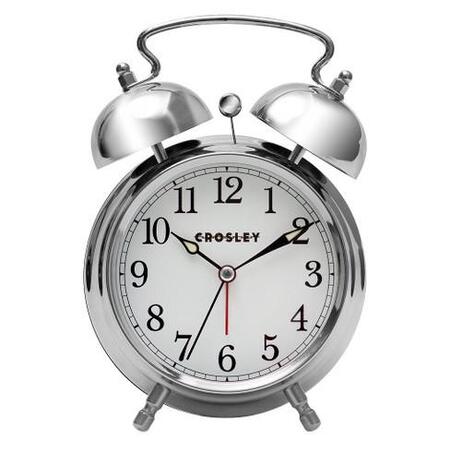}
        & \greenimg{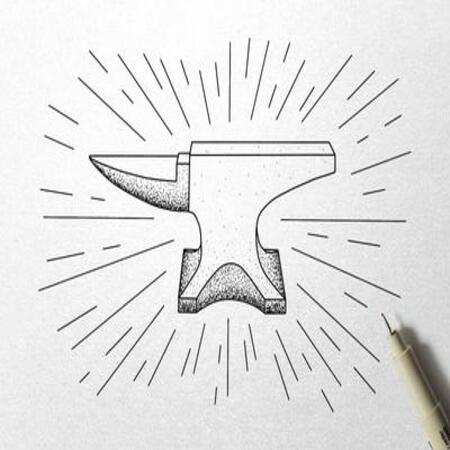} \\[1ex]
        ~ & \textsc{MNIST} & \textsc{SYN} & \textsc{SVHN}&\scriptsize \textsc{MNIST-M} & \textsc{USPS} 
          & \textsc{CLIPART} & \textsc{CLIPART} & \textsc{QUICKDRAW} & \textsc{REAL} & \textsc{SKETCH} \\[-0.2em]
        \raisebox{0.4cm}{\textbf{FreeSel}}
        & \greenimg{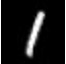}
        & \redimg{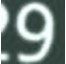}
        & \redimg{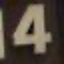}
        & \redimg{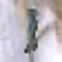}
        & \redimg{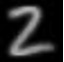}
        & \redimg{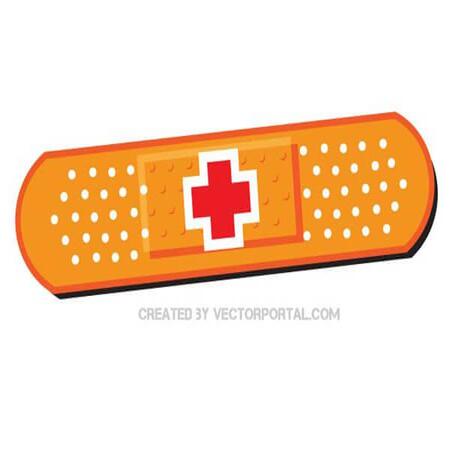}
        & \redimg{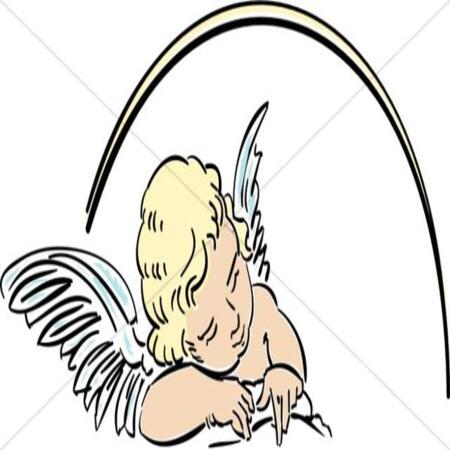}
        & \redimg{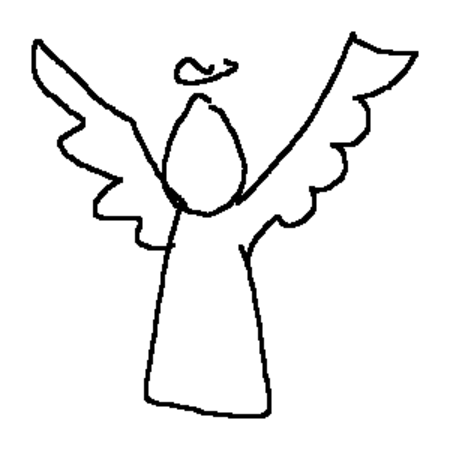}
        & \redimg{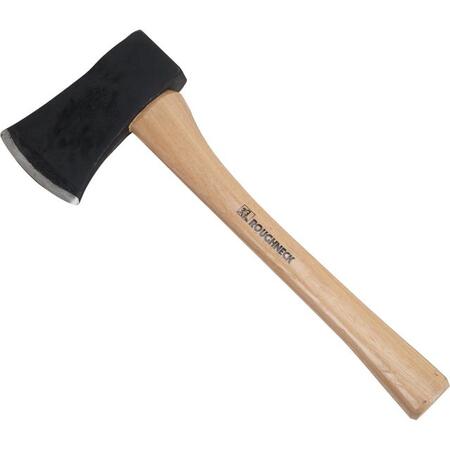}
        & \greenimg{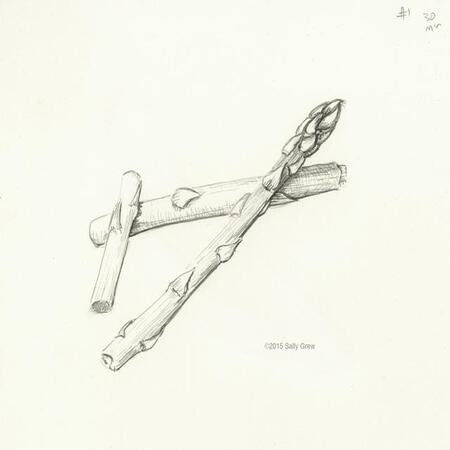}\\[1ex]
        ~ & \textsc{SYN} & \textsc{SYN} & \textsc{SYN} &\scriptsize \textsc{SYN} & \textsc{SYN}
          & \textsc{CLIPART} & \textsc{SKETCH}  & \textsc{QUICKDRAW} &  \textsc{CLIPART} &  \textsc{CLIPART} \\[-0.2em]
        \raisebox{0.4cm}{\textbf{ActiveFT }}
        & \redimg{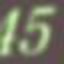}
        & \redimg{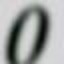}
        & \redimg{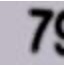}
        & \redimg{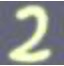}
        & \redimg{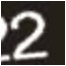}
        & \redimg{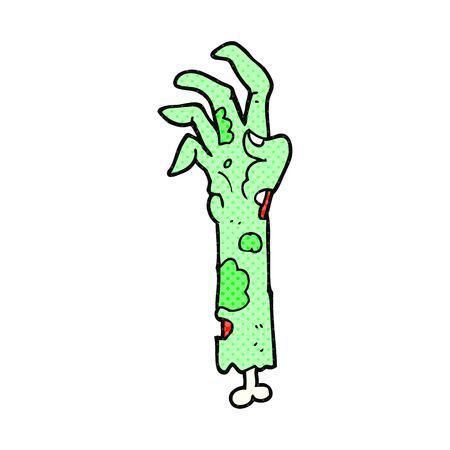}
        & \greenimg{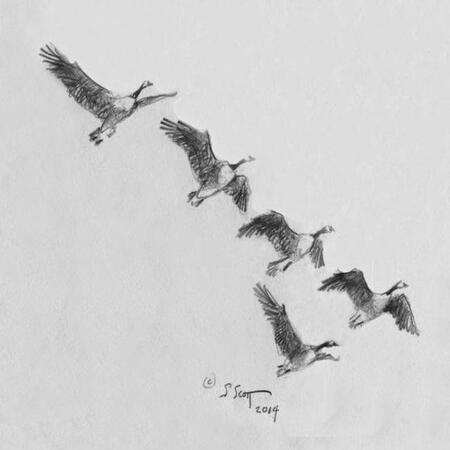}
        & \redimg{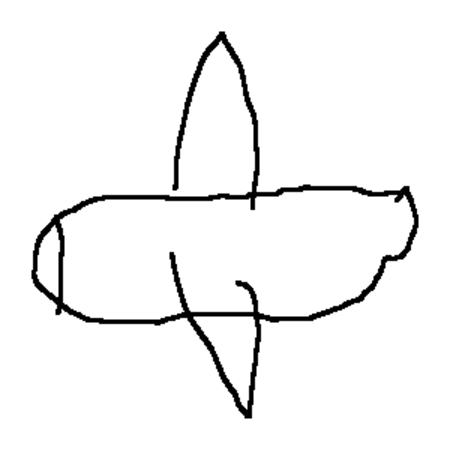}
        & \redimg{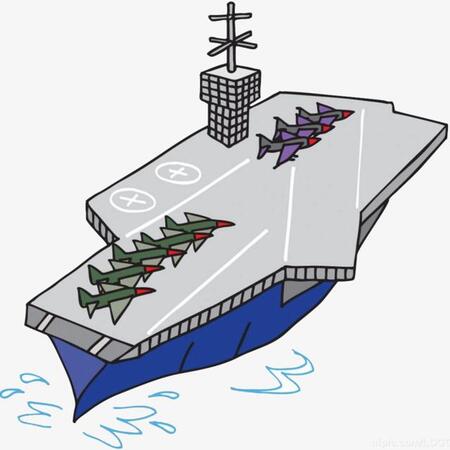}
        &\redimg{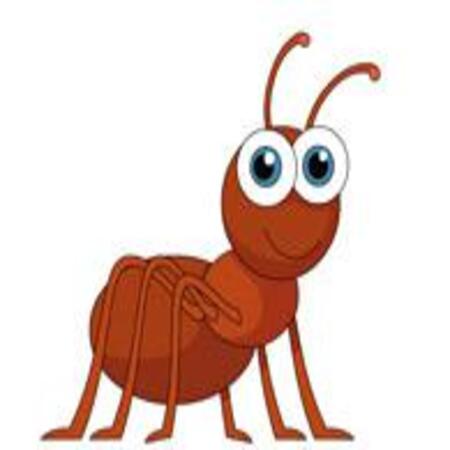}\\[1ex]
        ~ & \textsc{SYN} & \textsc{SYN} & \textsc{SYN} &\scriptsize \textsc{SYN} & \textsc{SYN}
          & \textsc{CLIPART} & \textsc{SKETCH}  & \textsc{CLIPART} &  \textsc{CLIPART}&  \textsc{REAL} \\[-0.2em]
        \raisebox{0.4cm}{\textbf{BiLAF}}
        & \redimg{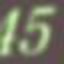}
        & \redimg{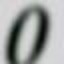}
        & \redimg{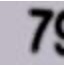}
        & \redimg{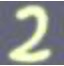}
        & \redimg{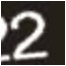}
        & \redimg{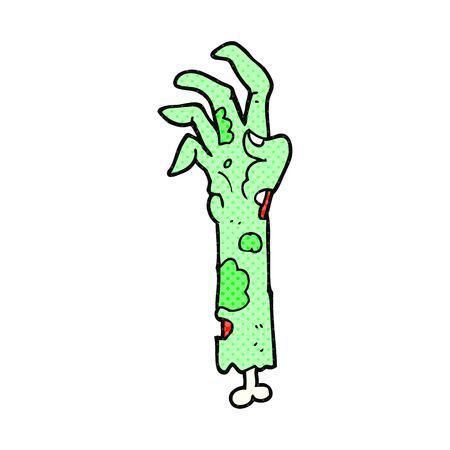}
        & \greenimg{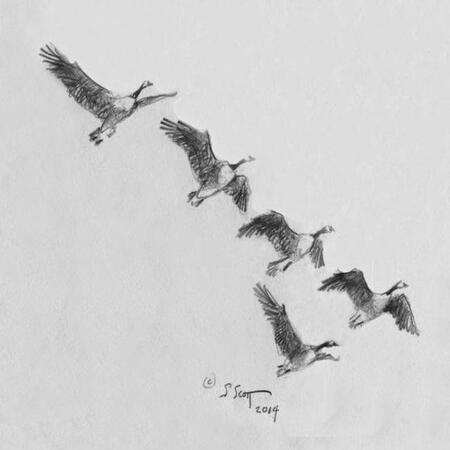}
        & \redimg{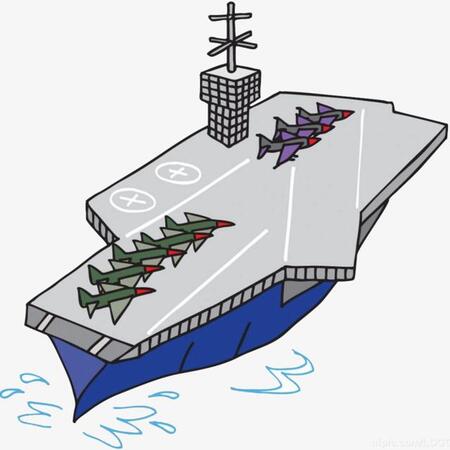}
        & \redimg{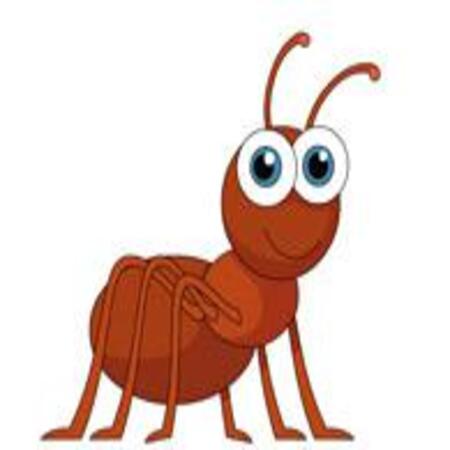}
        &\redimg{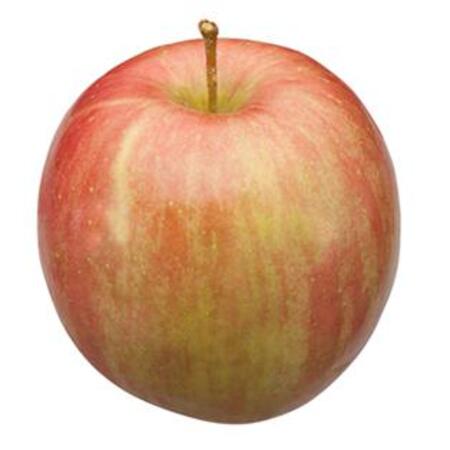}\\[1ex]
        ~ & \textsc{MNIST} & \textsc{MNIST} & \textsc{MNIST} &\scriptsize \textsc{MNIST} & \textsc{MNIST}
          & \textsc{SKETCH} & \textsc{SKETCH}  & \textsc{SKETCH} &  \textsc{SKETCH}&  \textsc{SKETCH} \\[-0.2em]
        \raisebox{0.4cm}{\textbf{   DaSH(ours)}}
        & \greenimg{Figures/images/qualitative_pic/DaSH_d_1_mnist.jpg}
        & \greenimg{Figures/images/qualitative_pic/DaSH_d_2_mnist.jpg}
        & \greenimg{Figures/images/qualitative_pic/DaSH_d_3_mnist.jpg}
        & \greenimg{Figures/images/qualitative_pic/DaSH_d_4_mnist.jpg}
        & \greenimg{Figures/images/qualitative_pic/DaSH_d_5_mnist.jpg}
        & \greenimg{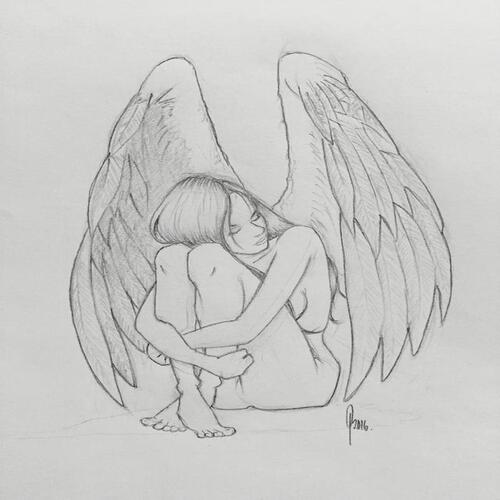}
        & \greenimg{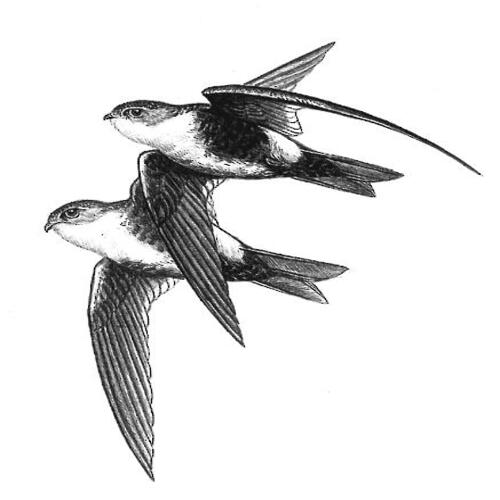}
        & \greenimg{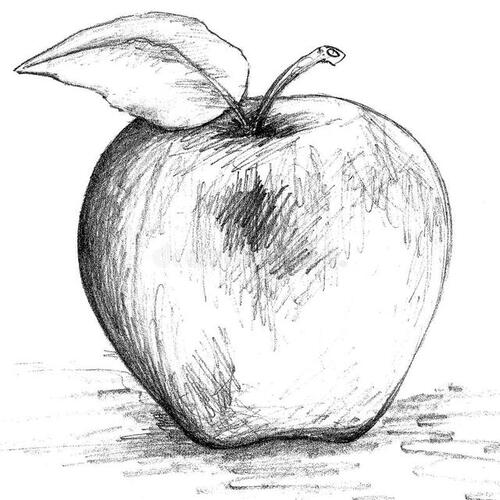}
        & \greenimg{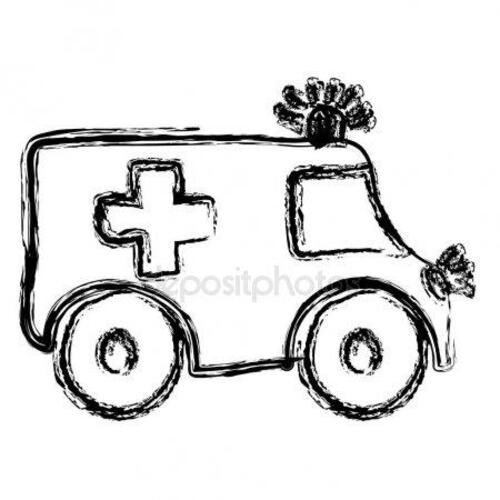}
        &\greenimg{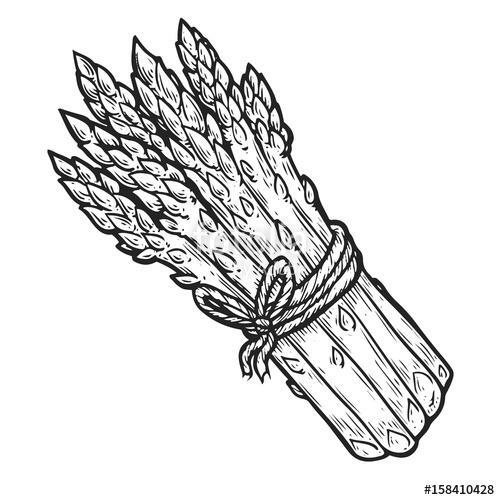}

    \end{tabular}}
    \caption{Qualitative comparisons on \textsc{Digit-Five} (target: \textsc{MNIST}) and \textsc{DomainNet} (target: \textsc{SKETCH}). Each selected image is labeled by its source domain (above), with green borders indicating a correct domain match to the target and red borders indicating a mismatch. Unlike prior methods, which frequently select subsets from mismatched domains in the first exploration step, \modelname{} consistently identifies subsets from the correct domain, even in challenging settings with visually similar categories.
    }
    \label{fig:qual_comp}
\end{figure*}

\begin{table}[t]
\centering
\resizebox{0.9\columnwidth}{!}{
\begin{tabular}{lcc}
\toprule
\textbf{Method} & \textbf{\# Steps} & \textbf{Accuracy} \\
\midrule
\dasflat{}                   & 163  & 90.9$\pm$2.0 \\
\modelname{}              & 140  & 91.2$\pm$0.9 \\
\modelname{} (cross-domain grouping)  & 154  & 92.2$\pm$0.7 \\
\bottomrule
\end{tabular}
}
\caption{Robustness of \modelname{} under cross-domain grouping. Performance on \textsc{USPS} with cross-domain groups, where each group contains exactly one dataset from each domain, removing opportunities to select multiple same-domain datasets. \modelname{} achieves the robust accuracy while requiring fewer steps than the non-hierarchical variant \dasflat{}.}
\label{tab:extreme_mixed}
\end{table}

\subsection{Robustness under Cross-Domain Grouping} 
We evaluate \modelname{} in an extreme cross-domain grouping scenario, where each group is constructed to contain exactly one dataset from each domain. This setup eliminates the possibility of selecting multiple same-domain datasets within a single group, stress-testing the ability of \modelname{} to perform effective selection when group structure does not align with domain semantics and offers no within-domain redundancy to exploit. As shown in Table 4, \modelname{} delivers robust accuracy and outperforms the non-hierarchical baseline, \dasflat{}, while also requiring fewer selection steps. Our ablation results consistently show that, under different settings, \modelname{} remains effective, maintaining strong performance with minimal computational overhead.

\section{Qualitative Analysis}
Figure~\ref{fig:qual_comp} illustrates clear qualitative differences in the selection behavior of each method. Green borders indicate that the selected data instance belongs to the target domain, while red borders indicate domain mismatches. Across both benchmarks, baseline methods such as Core-Sets, FreeSel, ActiveFT, and BiLAF often select subsets from visually similar but incorrect domains. For example, when \textsc{MNIST} is used as the local dataset, most baselines retrieve images that are visually distinct from the target domain. 
Only FreeSel selects a sample from \textsc{MNIST}, which is consistent with its relatively better quantitative performance (Table~\ref{tab:digit_five_accuracy}). The rest of the baselines fail to retrieve meaningful samples. 
In contrast, \modelname{} effectively selects relevant data. 
This behavior extends to \textsc{DomainNet}, where \modelname{} maintains domain-consistent selection across diverse categories. These results suggest that \modelname{} internalizes domain structure more effectively than prior methods, allowing it to identify relevant datasets even under distribution shift and candidate noise, an essential capability for transferability in collaborative data-sharing settings.

\section{Conclusion}\label{sec:conclusion}
This work addresses a key bottleneck in machine learning: selecting training datasets from diverse sources such as institutions, repositories, or collections. 
We introduce \modelname{}, a dataset selection framework that models the hierarchical relationship among datasets and data sources to improve selection efficiency and downstream performance. Experimental results demonstrate that \modelname{} consistently outperforms non-hierarchical and existing instance-level data selection baselines, and remains robust under realistic constraints such as imperfect grouping and limited exploration budgets. These findings underscore the importance of effectively automating practical data curation as machine learning models increasingly depend on large-scale heterogeneous data sources from various online repositories. Future directions include incorporating multi-objective selection criteria such as utility, fairness, and domain coverage, and applying \modelname{} to large-scale, multi-institutional data sharing platforms, where group membership and dataset availability evolve over time.

\section*{Acknowledgments}
This research is based on work partially supported by the National Science Foundation under award number CMMI-2331985, the U.S. Defense Advanced Research Projects Agency (DARPA) under award number HR001125C0303, and U.S. Army DEVCOM 
under award number W5170125CA160. The views and conclusions contained herein are those of the authors and should not be interpreted as representing the official policies, either expressed or implied, of NSF, DARPA, the U.S. Army, or the U.S. Government. The U.S. Government is authorized to reproduce and distribute reprints for governmental purposes notwithstanding any copyright annotation therein.

\bibliography{aaai2026}
\input{ReproducibilityChecklist}
\newpage
\clearpage
\newpage

\section{Datasets}\label{apd:data}

Our two experimental benchmarks are \textbf{\textsc{Digit-Five}} and \textbf{\textsc{DomainNet}}~\cite{peng2019moment}, both commonly employed to evaluate domain adaptation models \cite{schrod2023fact, yao2022federated, simon2022generalizing, jin2021re, komatsu2021multi, li2021dynamic, luo2021ensemble, singh2021clda}, with groupings based on different domain types for the same task. We briefly describe each below:

\paragraph{\textsc{Digit-Five}.} The \textsc{Digit-Five} dataset contains images of handwritten digits (0-9) with variability in writing styles, stroke thickness, and other characteristics. The dataset has five different data subsets: \textbf{\textsc{MNIST}} (clean, grayscale images of handwritten digits in a uniform style) \cite{lecun1998gradient}, \textbf{\textsc{MNIST-M}} (images from \textsc{MNIST} superimposed on complex color backgrounds from BSDS500) \cite{ganin2015unsupervised}, \textbf{\textsc{USPS}} (grayscale images of digits from scanned mail, with variations in scale and stroke thickness) \cite{hull1994database}, \textbf{\textsc{SVHN}} (real-world, full-color images of house numbers with diverse fonts and lighting conditions) \cite{roy1807effects}, and \textbf{\textsc{SYN}} (synthetic images of digits manipulated with various font styles and digital effects) \cite{ganin2015unsupervised}. For our experiments, we utilize preprocessed data as provided by \citet{schrod2023fact}.  
Images are encoded into vector representations using a CNN feature extractor with three convolutional layers followed by a pooling layer, trained on the dataset.
For each of the \textsc{Digit-Five} subsets, we randomly sample data points to divide them into three mutually exclusive groups. We refer to each \textsc{MNIST}-derived groups as {\{mn0, mn1, mn2\}}, groups derived from \textsc{MNIST-M} as {\{mm0, mm1, mm2\}}, from \textsc{USPS} as {\{us0, us1, us2\}}, from \textsc{SVHN} as {\{sv0, sv1, sv2\}}, and from \textsc{SYN} as {\{sy0, sy1, sy2\}}.

\paragraph{\textsc{DomainNet}.} The \textsc{DomainNet} dataset \cite{peng2019moment} contains data instances from diverse object categories across six domains: real, clipart, painting, sketch, infograph, and quickdraw, each representing a distinct style. For our experiments, we select images from 15 classes across four domains: \textbf{\textsc{CLIPART}} (clip art images), \textbf{\textsc{QUICKDRAW}} (drawings from the game “Quick Draw”), \textbf{\textsc{REAL}} (photos and real-world images), and \textbf{\textsc{SKETCH}} (sketches of objects).
Data pre-processing is similar to that for \textsc{Digit-Five}. For each of the domain subsets, we randomly sample data points to divide them into three mutually exclusive groups. 
We refer to groups from \textsc{CLIPART} as {\{cp0, cp1, cp2\}}, \textsc{QUICKDRAW} groups as {\{qd0, qd1, qd2\}}, groups from \textsc{REAL} as {\{rl0, rl1, rl2\}}, and groups from \textsc{SKETCH} as {\{sk0, sk1, sk2}\}.

To assess \modelname{}'s robustness under different group configurations, we experiment with three distinct settings:

\noindent\textbf{Perfect Grouping:} Here, groups have clear domain boundaries, each containing three distinct datasets: \{mn0, mn1, mn2\}, \{mm0, mm1, mm2\}, \{us0, us1, us2\}, \{sv0, sv1, sv2\}, and \{sy0, sy1, sy2\}. Similarly, \textsc{DomainNet} is partitioned into coherent domain-aligned groups: \{cp0, cp1, cp2\}, \{qd0, qd1, q2\}, \{rl0, rl1, rl2\}, and \{sk0, sk1, sk2\}.

\noindent\textbf{Mixed Grouping:} We consider mixed groups that contain subsets from different domains. This reflects real-world situations where organizations may contribute data spanning multiple domains. For \textsc{Digit-Five}, we define the following groups: \{mn1, mn2, mm0\}, \{mm1, mm2, us0\}, \{us1, us2, sv0\}, \{sv1, sv2, sy0\}, \{sy1, sy2, mn0\}. For \textsc{DomainNet}, the groups are: \{cp1, cp2, qd0\}, \{qd1, qd2, rl0\}, \{rl1, rl2, sk0\}, \{sk1, sk2, cp0\}.

\noindent\textbf{Cross-Domain Grouping:} We construct groups such that no group contains datasets from the same domain. This tests whether the method can still make effective selections when group structure does not reflect underlying domain similarity. The \textsc{Digit-Five} groups are: \{mn0, sv0, mm0\}, \{sv1, mm1, us0\}, \{mm2, us1, sy0\}, \{us2, sy1, mn1\}, \{sy2, mn2, sv2\}.

\section{Implementation details}\label{apd:setup}
For the \textsc{Digit-Five} dataset, the local classifiers consist of a single CNN layer. 
Figure \ref{fig:gt_acc} (a) shows the ground truth accuracy heatmap for \textsc{Digit-Five}, where the first column displays the local accuracy (loc) for each digit classifier on the 
\textsc{MNIST} ({\{mn0, mn1, mn2\}}), \textsc{MNIST-M} ({\{mm0, mm1, mm2\}}), \textsc{USPS} ({\{us0, us1, us2\}}), \textsc{SVHN} ({\{sv0, sv1, sv2\}}) and \textsc{SYN} ({\{sy0, sy1, sy2\}}) subgroups while the last column reveals the global accuracy achieved after each classifier is trained on the relevant subsets sampled from its corresponding dataset. For example, the global accuracy of 89.3\% for \textsc{MNIST} is achieved by training the local model on {\{mn0, mn1, mn2\}}. Training on other datasets yields lower accuracy than the local accuracy, suggesting a degradation in performance. Therefore, the optimal performance for \textsc{MNIST} is attained by training on {\{mn0, mn1, mn2\}}. The middle columns depict accuracy of local classifiers after additional training on each individual subset.
For the \textsc{DomainNet}\cite{peng2019moment} dataset, the local classifier consists of three fully connected layers. 
Figure \ref{fig:gt_acc} (b) shows that the \textsc{CLIPART} model exhibits the lowest local accuracy at 40.7\%, while the sketch model achieves the highest local accuracy at 67\%. In this benchmark, although the local model still gains the most improvement when trained on external sets from the same domain,  datasets from other domains also improve model accuracy. For instance, \textsc{CLIPART} datasets \{cp0, cp1, cp2\} contribute to enhancing local model performance for the \textsc{REAL} dataset group. These characteristics render the \textsc{DomainNet} experiments closer to realistic data-sharing settings.

The non-hierarchical baseline, \dasflat{}, serves as a flat \modelname{} variant to directly compare the utility of hierarchical decomposition. \dasflat{} treats each dataset independently without modeling shared origin or source-level relationships. 
For \modelname{} and \dasflat{}, we set $\mu_0$ and $\theta_i$ to 0, and $\sigma_0^2$ and $\hat{\sigma}_0^2$ to 2, as prior distributions for all dataset groups and datasets. We set the pre-defined percentile posterior mean threshold to 80 and 60 for the perfect and mixed groups, respectively.
At every time step, \modelname{} decides on a dataset to select, retrieves a sample, and the local model predicts the sample's label. The accuracy of this prediction determines the reward, i.e., \( r_{i,j}(t) = 1 \) if \( \hat{y} = y \), and \( r_{i,j}(t) = 0 \) otherwise, where \( \hat{y} \) represents the predicted label and \( y \) the actual label of the sample. This reward, either 1 for a correct prediction or 0 for an incorrect one, serves as the sole feedback for the algorithm to update its prior beliefs. \modelname{} systematically refines these beliefs in response to the observed reward outcomes. 

For accurate and efficient dataset selection, we employ K-means clustering to identify representative data points, selecting five points nearest to the centroids in each cluster to encapsulate the dataset's characteristics. Specifically, for \textsc{Digit-Five}, which comprises 10 distinct classes, we configure clustering to generate 10 clusters to ensure that the variability inherent in each class is captured effectively. The model's priors are updated exclusively using 5 near-centroid points from each cluster. Similarly, for \textsc{DomainNet}, we generate 15 clusters corresponding to the 15 classes in the dataset and use 5 near-centroid points from each cluster.

Data selection stops when all representative points from a particular dataset are selected, indicating that the selection model has identified a specific dataset as likely to significantly enhance model performance. The total number of steps required to explore all representative points from all 15 \textsc{Digit-Five} data subsets is 750 (corresponding to the 15 data subsets, each with 10 clusters and 5 near-centroid points for each cluster). Similarly, the total number of steps required to explore all representative points from \textsc{DomainNet} is 1125. However, our experiments verify that the proposed empirical stopping criterion requires significantly fewer steps.

\subsection{\modelname{} Dataset Selection Algorithm}
We include the full pseudocode of the proposed \modelname{} dataset selection algorithm in Algorithm~\ref{HB}, capturing the hierarchical selection process over groups and datasets. The pseudocode corresponds to the framework described in Section~\ref{sec:methods}

\begin{algorithm}[t!]
\caption{\modelname{} Dataset Selection}
\label{HB}
\begin{algorithmic}[1]  
\STATE Initialize $P(\theta_i|r_i)$ group distributions, $P(\theta_{i,j}|r_{i,j})$ dataset distributions, and $\mathcal{N}(\theta_{i,j}, \sigma_r^2)$ reward distributions
\FOR{$t = 1,\dots,T$}
    \FOR{$i = 1,\dots,n$}
        \STATE Sample $\hat{\theta}_{i}(t) \sim P(\theta_i|r_i)$
    \ENDFOR
    \STATE $i = \arg\max \{\hat{\theta}_{i}(t) \mid i \in [n]\}$ \COMMENT{$g_i$ is chosen}
    \FOR{$j = 1,\dots,m$}
        \STATE Sample $\hat{\theta}_{i,j}(t) \sim P(\theta_{i,j}|r_{i,j})$
    \ENDFOR
    \STATE $j = \arg\max \{\hat{\theta}_{i,j}(t) \mid j \in [m]\}$ \COMMENT{$d_{i,j}$ is chosen}
    \STATE Receive reward $r_{i,j}(t) = \mathbf{1}\{\hat{y} = y\}$
    \STATE Update $P(\theta_i|r_i)$ and $P(\theta_{i,j}|r_{i,j})$ \COMMENT{Eq.~(\ref{eq:2}) and Eq.~(\ref{eq:3})}
\ENDFOR
\end{algorithmic}
\end{algorithm}
\subsection{Scalability to Larger Dataset Pools}
\begin{figure}[t!]
    \centering
    \includegraphics[width=\columnwidth]{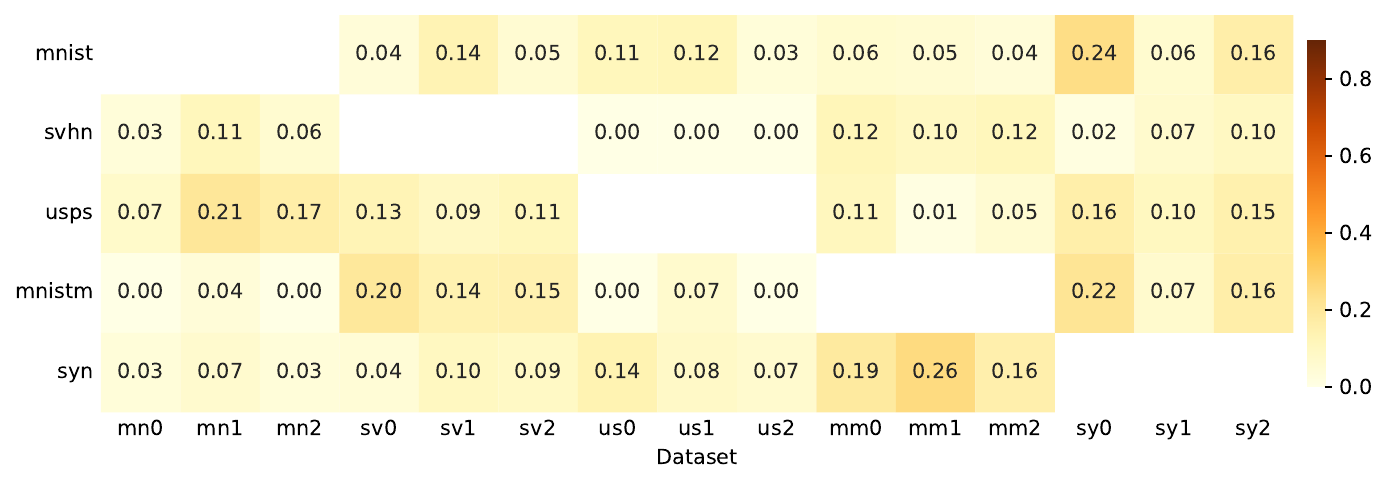}
   \caption{\textbf{\modelname{} reliably signals the absence of relevant datasets.} When no beneficial datasets are present in the pool, the posterior means remain consistently low, even after 600 exploration steps, indicating that \modelname{} does not overcommit to low-utility sources.}
    \label{fig:non_relevant}
\end{figure}

We evaluate the scalability of \modelname{} by expanding the number of candidate datasets within each \textsc{Digit-Five} group. Specifically, the \textsc{MNIST}, \textsc{SVHN}, \textsc{USPS}, \textsc{MNIST-M}, and \textsc{SYN} groups are augmented to include 10, 12, 11, 9, and 9 datasets, respectively. As shown in Table~\ref{tab:digitfive_scalability}, \modelname{} continues to identify high-utility datasets and consistently improves downstream accuracy across all domains. Importantly, per-step computational cost remains constant, and the total number of selection steps increases sublinearly with the size of the dataset pool. For instance, \textsc{SVHN} contains 4× more datasets than in the original setting, yet \modelname{} requires only 2.6× more steps. These results highlight the method’s scalability and efficiency in more complex selection settings.

\begin{table}[t!] 
\centering
\caption{\textbf{\modelname{} scalability ablation on \textsc{Digit-Five}.} Accuracy across five domains with 15 vs. 51 dataset configurations. Larger dataset pools improve performance consistently.}
\label{tab:digitfive_scalability}
\resizebox{\columnwidth}{!}{
\begin{tabular}{@{}lccccccl@{}}
\toprule
\textbf{Domain} & \textbf{\textsc{MNIST}} & \textbf{\textsc{SVHN}} & \textbf{\textsc{USPS}} & \textbf{\textsc{MNIST-M}} & \textbf{\textsc{SYN}} & \textbf{AVG} \\ 
\midrule
\modelname{} (15) & 89.5 & 69.2 & 91.2 & 78.9 & 62.9 & 78.3 \\
\modelname{} (51) & \textbf{93.7} & \textbf{71.4} & \textbf{92.8} & \textbf{83.4} & \textbf{76.5} & \textbf{83.6} \\
\bottomrule
\end{tabular}
}
\end{table}

\subsection{Robustness to Absence of Relevant Sources} As in real-world applications where the usefulness of datasets is not known in advance, we further evaluate the behavior of \modelname{} when no relevant datasets are available in the candidate pool. As illustrated in Figure~\ref{fig:non_relevant}, in this setting, \modelname{} continues exploration as instructed, but the inferred posterior means across all datasets remain consistently low. This indicates that the method robustly recognizes the lack of beneficial datasets, producing a clear signal that no source meaningfully improves downstream performance. \modelname{} avoids committing to low-utility datasets, demonstrating reliable behavior even in unfavorable selection conditions.

\subsection{Optimality Analysis}
Let $\theta^\star=\max_{d_{i,j}\in\mathcal{D}}\theta_{i,j}$ denote the optimal expected reward and fix
\[
d^\star \in \arg\max_{d_{i,j}\in\mathcal{D}} \theta_{i,j}.
\]
Define the sub–optimality gaps
\[
\Delta_{i,j}=\theta^\star-\theta_{i,j}>0
\qquad\text{for all } d_{i,j}\neq d^\star.
\]
Under the hierarchical model, \modelname{} maintains posterior distributions for both the group-level parameters $\theta_i$ and the dataset-level parameters $\theta_{i,j}$.  
Under the Gaussian posterior updates in Eqs.~(\ref{eq:3}) and~(\ref{eq:4}), the corresponding posterior variances 
$\lambda_i^2(t)$ and $\lambda_{i,j}^2(t)$ satisfy 
\[
\lambda_i^2(t)\to 0,\qquad \lambda_{i,j}^2(t)\to 0,
\]
so the posterior distributions concentrate around each true $\theta_i$ and $\theta_{i,j}$.   
Thus the probability of selecting a sub–optimal dataset vanishes, and
\[
\Pr\!\big(D(t)=d^\star\big)\longrightarrow 1 \quad \text{as } t\to\infty.
\]

Let $R_T=\sum_{t=1}^T(\theta^\star-\theta_{D(t)})$ be the cumulative regret.  
Standard results for Gaussian Thompson Sampling imply that each sub–optimal dataset $d_{i,j}$ is selected in expectation 
$O\!\left((\sigma_r^2/\Delta_{i,j}^2)\log T\right)$ times.  
Therefore,
\[
R_T
=O\!\left(
\sum_{d_{i,j}\neq d^\star}
\frac{\sigma_r^2}{\Delta_{i,j}}\,\log T
\right),
\]
which achieves asymptotically optimal $O(\log T)$ regret.

\subsection{Limitations} 
While \modelname{} is designed for settings where data is organized into discrete, variably relevant datasets grouped by source or origin, our evaluation has focused on publicly available image datasets. This allows for controlled benchmarking but may not capture the full complexity of other data modalities or selection environments. In future work, we plan to extend \modelname{} to additional domains such as time-series and graph data, where hierarchical structure may arise from different sensors, sources, or collection protocols.

\subsection{Broader Impacts}
By modeling selection hierarchically, \modelname{} improves accuracy–efficiency trade-offs in realistic deployment scenarios. As such methods become more common, they can streamline large-scale data integration and expand access to high-quality training data. By making dataset selection more robust, efficient, and transparent, \modelname{} supports ML systems that reflect the constraints and diversity of real-world data ecosystems.

\end{document}